%% file: main.tex
\documentclass[conference]{IEEEtran}
\IEEEoverridecommandlockouts
\usepackage{cite}
\usepackage{amsmath,amssymb,amsfonts}
\usepackage{graphicx}
\usepackage{textcomp}
\usepackage{xcolor}

\usepackage{subcaption}
\usepackage{multirow}
\usepackage{comment}
\usepackage{algpseudocode}
\usepackage{algorithm}

\def\BibTeX{{\rm B\kern-.05em{\sc i\kern-.025em b}\kern-.08em
    T\kern-.1667em\lower.7ex\hbox{E}\kern-.125emX}}
\begin{document}

\def\cblue{\textcolor{blue}}
\def\cred{\textcolor{red}}
\def\cmag{\textcolor{magenta}}
\def\cgreen{\textcolor{black}}

\title{On the Impact of Sampling on Deep \\ Sequential State Estimation
\thanks{This work has been partially supported by the NSF under Awards ECCS-1845833 and CCF-2326559.}
}

\author{Helena Calatrava$^*$, Ricardo~Augusto~Borsoi$^\dag$, Tales Imbiriba$^{*,\ddag}$, Pau Closas$^*$ \bigskip\\
\normalsize 
$^*$ Electrical and Computer Engineering Dept., Northeastern University, Boston, MA, USA \\
$^\dag$ CRAN, University of Lorraine, CNRS, Vandoeuvre-les-Nancy, France \\
$^\ddag$ Institute for Experiential AI, Northeastern University, Boston, MA, USA \\
calatrava.h@northeastern.edu, ricardo-augusto.borsoi@cnrs.fr, t.imbiriba@northeastern.edu, closas@northeastern.edu
}



\maketitle

\begin{abstract}
\input{paper_sections/abstract}
\end{abstract}


\section{Introduction}
\label{sec:intro}
\input{paper_sections/intro}

\section{System Model}
\label{sec:system_model}
\input{paper_sections/system_model}
%
%
\cgreen{\section{Importance Weighted Deep Kalman Filter}} 
\label{sec:IS_for_DKF}
\input{paper_sections/method}
%


\section{Experiments}
\label{sec:experiments}
\input{paper_sections/experiments}

\section{Conclusion}
\label{sec:conclusion}
\input{paper_sections/conclusion}


\bibliographystyle{IEEEtran}
\bibliography{IS_VI}

\end{document}

%% file: paper_sections/abstract.tex
State inference and parameter learning in sequential models can be successfully performed with approximation techniques that maximize the evidence lower bound (ELBO) to the marginal log-likelihood of the data distribution. These methods may be referred to as Dynamical Variational Autoencoders (DVAEs), and our specific focus lies on the deep Kalman filter (DKF).
It has been shown that the ELBO objective can oversimplify data representations, potentially compromising estimation quality.
Tighter Monte Carlo objectives (MCOs) have been proposed in the literature to enhance generative modeling performance. For instance, the importance weighted autoencoder (IWAE) objective uses importance weights to reduce the variance of marginal log-likelihood estimates.
In this paper, sampling is applied to the DKF framework for learning deep Markov models (DMMs), resulting in the importance weighted DKF (IW-DKF), which shows an improvement in terms of log-likelihood estimates and Kullback–Leibler (KL) divergence between the variational distribution and the transition model. 
The framework using the sampled DKF update rule is also accommodated to address sequential state and parameter estimation when working with highly non-linear physics-based models. An experiment with the 3-space Lorenz attractor shows an enhanced generative modeling performance and also a decrease in RMSE when estimating the model parameters and latent states, indicating that tighter MCOs lead to improved state inference performance.
%

%% file: paper_sections/intro.tex
%

Most methodologies for training deep generative models require posterior inference. However, the recovery of the latent variables is not the primary objective but rather an inherent outcome of the model training process.
Variational autoencoders (VAEs)~\cite{kingma_auto-encoding_2022, rezende_stochastic_2014} provide a framework for learning the parameters of flexible generative models (the \textit{decoder)} while approximating their intractable posterior with a parametric inference model (the \textit{encoder}).
%
The model is learned by maximizing the evidence lower bound (ELBO) to the marginal log-likelihood of the data distribution, which is also intractable. 
%
%
The encoder output is usually assumed to be a lower-dimensional and unobserved latent variable that generates higher-dimensional observed random variables through a probabilistic process~\cite{dynamical_vaes}. 
%
%
%
%
The concept of combining inference and generative model learning becomes more intuitive when viewed through the lens of the Expectation-Maximization (EM) algorithm~\cite{1977_em_algo}, which alternatingly optimizes the ELBO with respect to the generative model parameters, and with respect to the variaitonal posterior. This perspective is well explained in, e.g.,~\cite{bishop,Neal1998}.
%

%
While VAEs are often of interest due to their unsupervised representation learning capabilities for data generation and reconstruction, they also offer efficient inference and parameter estimation within Bayesian frameworks. This is especially interesting when modeling sequential data that exhibits temporal correlation, since the temporal structure can be exploited within this framework. Several works in the literature aim at learning dynamical models while estimating meaningful continuous latent variables~\cite{kalman_variational_autoencoders, recurrent_variational_autoencoders}. In~\cite{dynamical_vaes}, a thorough overview of these models, which are referred to as dynamical VAEs (DVAEs), is presented. In this paper, our primary focus centers on the deep Kalman filter (DKF), which was originally proposed in~\cite{krishnan_deep_2015} and further elaborated in~\cite{krishnan_structured_2016}. The DKF is a fusion of state space models (SSMs) and deep neural networks, as seen in works such as~\cite{haykin_2001}, while also incorporating the VAE learning framework.

%

Although maximizing the lower bound defined by the ELBO is a widely used objective~\cite{blei_variational_2017}, it may lead to a simplified representation of the data, failing to use the entire modeling capacity of the generative and inference models. Considering this, efforts have been made in the literature to introduce tighter Monte Carlo objectives (MCOs) that compromise estimation quality to a lesser extent.
MCOs generalize the ELBO to any objective function defined by taking the logarithm of a positive, unbiased estimator of the data marginal log-likelihood. 
The work in~\cite{maddison_filtering_2017} shows that the tightness of an MCO scales like the relative variance of the estimator from which it is built. 
For instance, the importance weighted autoencoder (IWAE)~\cite{burda_importance_2016} shares the VAE architecture but is trained on a tighter lower bound derived from the $K$-sample importance weighting estimate of the marginal log-likelihood, thus approaching the true value for increasing $K$.
%
Importance weighting methods facilitate the approximation of the intractable true posterior distribution, which can often only be evaluated up to the normalizing constant~\cite{bugallo_adaptive_2017}.

%

%
For models with sequential structure, the variance of the likelihood estimator of particle filters scales more favorably than importance sampling.
Thus, a family of particle-based MCOs that could serve as tighter objectives than the IWAE was suggested in~\cite{maddison_filtering_2017}. These were referred to as \textit{filtering variational objectives}. 
%
Overall, there is a significant amount of research focused on designing tighter MCOs that provide a lower bound that is closer to the data. However, these works focus on its impact on generative modeling, while the impact of a tighter MCO on parameter learning and state inference is less clear.
This work investigates how a tighter bound influences the performance of the inference model when working with sequential data. 

%

%
%

In this paper, we propose the importance weighted DKF (IW-DKF), which uses sampling as in the IWAE to improve state inference performance in sequential models. We adapt the VAE update rule by applying the $K$-sample importance weighting estimate of the marginal log-likelihood from the IWAE and extending it to the temporal setting that characterizes a general DVAE. Specifically, the inference network is structured by maintaining the conditional independence and Markovity assumptions in the model.
Considering that the IWAE provides higher log-likelihoods on density estimation benchmarks when working with non-sequential data, our hypothesis is that applying importance sampling to the DKF objective function may enhance state inference performance in complex sequential models.
This is evaluated in two experiments. In the first experiment, a deep Markov model (DMM) is learned with the IW-DKF objective, using the regular DKF objective as a benchmark. In the second experiment, the framework is accommodated to provide efficient state and parameter estimation when working with highly non-linear physics-based models, specifically a 3-space Lorenz attractor model. Results indicate that tighter MCOs can improve both the latent variable inference and the parameter estimation performance in this setting.

%% file: paper_sections/system_model.tex
%
State estimation allows us to find estimates of the vector valued time series $\mathbf{z}_{1:T}$, with $\mathbf{z}_t \in \mathbb{R}^{n_z}$ and $\mathbf{z}_{a:b} = \{\mathbf{z}_a,\hdots,\mathbf{z}_b\}$, on the basis of a set of vector valued noisy observations $\mathbf{x}_{1:T}$, with $\mathbf{x}_t \in \mathbb{R}^{n_x}$~\cite{dunik2020state}.
The latent states prior, transition and emission models are given by
%
%
%
\begin{gather} 
    \mathbf{z}_1 \sim  \mathcal{N}\big(\mathbf{z}_1\, |\, \boldsymbol{\mu}_0,\boldsymbol{\Sigma}_0\big) \\
    \mathbf{z}_t\, |\, \mathbf{z}_{t-1} \sim  \mathcal{N}\big(\mathbf{z}_t\, |\, \mathbf{F}_{\boldsymbol\alpha}(\mathbf{z}_{t-1}), \mathbf{Q}_{\boldsymbol\gamma}(\mathbf{z}_{t-1})\big) \\
     \mathbf{x}_t\, |\, \mathbf{z}_t \sim \mathcal{N}\big(\mathbf{x}_t\, |\, \mathbf{H}_{\boldsymbol\kappa}(\mathbf{z}_t), \mathbf{R}_{\boldsymbol\lambda}(\mathbf{z}_t)\big) \,.
\end{gather}
%

The parameters of this generative model are collectively denoted within the vector ${\boldsymbol{\theta}}=(\boldsymbol\alpha,\,\boldsymbol\gamma,\,\boldsymbol\kappa,\,\boldsymbol\lambda)^\top$. 
%
%
The distributions of the latent states and observations are assumed to be conditionally Gaussian with differentiable mean and covariance functions.
%
%
%
We assume a first-order Markov model, meaning that states conditioned on the past state $\mathbf{z}_{t-1}$ are independent of other states or observations as $p_{\boldsymbol{\theta}}(\mathbf{z}_t|\mathbf{z}_{1:t-1}, \mathbf{x}_{1:t-1})=p_{\boldsymbol{\theta}}(\mathbf{z}_t|\mathbf{z}_{t-1})$. Additionally, the conditional independence of measurements implies that observations conditioned on the current state $\mathbf{z}_t$ do not depend on previous states or observations as $p_{\boldsymbol{\theta}}(\mathbf{x}_t|\mathbf{z}_{1:t}, \mathbf{x}_{1:t-1})=p_{\boldsymbol{\theta}}(\mathbf{x}_t|\mathbf{z}_t)$.
As a result, the mean $\mathbf{F}_{\boldsymbol\alpha}$ and covariance $\mathbf{Q}_{\boldsymbol\gamma}$ are functions of the previous latent state, whereas $\mathbf{H}_{\boldsymbol\kappa}$ and $\mathbf{R}_{\boldsymbol\lambda}$ depend on the current latent state.
With these equations, a large family of linear and non-linear Gaussian SSMs (GSSMs) is encompassed.
%
A particular case of GSSMs is DMMs, for which the parametric form of the model is unknown~\cite{krishnan_structured_2016}. With DMMs, the emission and transition models are replaced with complex multi-layer perceptrons. This leverages the representational power of deep neural networks to model complex high dimensional data, all while preserving the underlying Markovian structure of a hidden Markov model.
%

%% file: paper_sections/method.tex


\subsection{Dynamical Variational Autoencoders}
The joint distribution of the considered generative model can be factorized as $p_{\boldsymbol{\theta}}(\mathbf{x},\mathbf{z}) = p_{\boldsymbol{\theta}}(\mathbf{z})p_{\boldsymbol{\theta}}(\mathbf{x}|\mathbf{z})$. For the sake of notation clarity, we define $\mathbf{x} = \mathbf{x}_{1:T}$ and $\mathbf{z} = \mathbf{z}_{1:T}$. The so-called inference network $q_{\boldsymbol{\phi}}(\mathbf{z}|\mathbf{x})$ is a parametric conditional distribution with parameters ${\boldsymbol{\phi}}$ that approximates the intractable posterior distribution $p_{\boldsymbol{\theta}}(\mathbf{z}|\mathbf{x})$. 
The ELBO is derived by applying Jensen's inequality to the marginal log-likelihood as
\begin{align}
\label{eq:elbo1d}
    \begin{split}
  &\log p_{\boldsymbol{\theta}}(\mathbf{x}) 
  \geq 
  \mathop{\mathbb{E}}_{q_{\boldsymbol{\phi}}(\mathbf{z}|\mathbf{x})}\left[\log\frac{p_{\boldsymbol{\theta}}(\mathbf{x},\mathbf{z})}{q_{\boldsymbol{\phi}}(\mathbf{z}|\mathbf{x})}\right]  
  = 
\mathcal{L}^{\text{ELBO}}(\boldsymbol{\theta}, \boldsymbol{\phi};\mathbf{x}) \\& = 
\mathop{\mathbb{E}}_{q_{\boldsymbol{\phi}}(\mathbf{z}|\mathbf{x})}\left[\log p_{\boldsymbol{\theta}}(\mathbf{x}|\mathbf{z})\right] - \text{KL}(q_{\boldsymbol{\phi}}(\mathbf{z}|\mathbf{x})||p_{\boldsymbol{\theta}}(\mathbf{z}))\,.
\end{split}
\end{align}
The ELBO is maximized with respect to both $\boldsymbol\phi$ and $\boldsymbol\theta$ to learn the generative model and the inference network. This maximization can be performed efficiently using gradient-based methods and the reparameterization trick~\cite{kingma_auto-encoding_2022, kingma_introduction_2019}.
This bound depends on the Kullback–Leibler (KL) divergence between the variational posterior and the prior of the latent states.
%
%
%
%
%
%
%
Considering the temporal setting which characterizes the DVAEs, the expectation in~\eqref{eq:elbo1d} may be factorized as
\begin{align}
\label{eq:elbo_dvae}
    \begin{split}
  &\mathcal{L}^{\text{DVAE}}({\boldsymbol{\theta}}, {\boldsymbol{\phi}}; \mathbf{x}) = 
  \sum^{T}_{t=1} \mathop{\mathbb{E}}_{q_{\boldsymbol{\phi}}(\mathbf{z}_{1:t}|\mathbf{x})}\big[\log p_{\boldsymbol{\theta}}\left(\mathbf{x}_t|\mathbf{x}_{1:t-1}, \mathbf{z}_{1:t}\right)\big] 
  \\ 
  & \hspace*{-0.2cm} - \sum^{T}_{t=1} \mathop{\mathbb{E}}_{q_{\boldsymbol{\phi}}(\mathbf{z}_{1:t-1}|\mathbf{x})}\big[\text{KL}\left(q_{\boldsymbol{\phi}}(\mathbf{z}_t|\mathbf{z}_{1:t-1},\mathbf{x}\right)||p_{\boldsymbol{\theta}}(\mathbf{z}_t|\mathbf{x}_{1:t-1},\mathbf{z}_{1:t-1}))\big].
\end{split}
\end{align}
\normalsize
\subsection{Deep Kalman Filter}
\label{sec:sub:dkf}
In the particular case of the DKF, the conditional independence of measurements and a first-order Markov model are assumed, allowing the objective function to be written as
\begin{align}
\label{eq:elbo_dkf}
    \begin{split}
  &\mathcal{L}^{\text{DKF}}({\boldsymbol{\theta}}, {\boldsymbol{\phi}}; \mathbf{x}) = 
  \sum^{T}_{t=1} \mathop{\mathbb{E}}_{q_{\boldsymbol{\phi}}(\mathbf{z}_{t}|\mathbf{x})}\big[\log p_{\boldsymbol{\theta}}\left(\mathbf{x}_t|\mathbf{z}_t\right)\big]
  \\ & - \text{KL}(q_{\boldsymbol{\phi}}(\mathbf{z}_1|\mathbf{x})||p_0(\mathbf{z}_1))
  \\ & - \sum^{T}_{t=2} \mathop{\mathbb{E}}_{q_{\boldsymbol{\phi}}(\mathbf{z}_{t-1}|\mathbf{x})}\big[\text{KL}\left(q_{\boldsymbol{\phi}}(\mathbf{z}_t|\mathbf{z}_{t-1},\mathbf{x}\right)||p_0(\mathbf{z}_t|\mathbf{z}_{t-1}))\big] \,.
\end{split}
\end{align}

Let us consider 
a minibatch $\mathcal{D}^N = \{\mathbf{x}^{(i)}\}^{N}_{i=1}$ including $N$ i.i.d. datapoints corresponding to samples of the random variable $\mathbf{x}$.
The expectation in~\eqref{eq:elbo_dkf} can be approximated using Monte Carlo integration, leading to the DKF approximated objective function for the datapoint $\mathbf{x}^{(i)}$ as
\begin{align}
\label{eq:mc_approx_dkf}
    \begin{split}
  &\tilde{\mathcal{L}}^{\text{DKF}}({\boldsymbol{\theta}}, {\boldsymbol{\phi}}; \mathbf{x}^{(i)}) = 
  \frac1{L}\sum_{\ell=1}^{L}\sum^{T}_{t=1} \log p_{\boldsymbol{\theta}}(\mathbf{x}_t^{(i)}|\mathbf{z}_t^{(\ell)} )
  \\ & 
  - \text{KL}(q_{\boldsymbol{\phi}}(\mathbf{z}_1|\mathbf{x}^{(i)})||p_0(\mathbf{z}_1))
  \\ & 
  - \frac1{L}\sum_{\ell=1}^{L}\sum^{T}_{t=2}\text{KL}(q_{\boldsymbol{\phi}}(\mathbf{z}_t|\mathbf{z}_{t-1}^{(\ell)},\mathbf{x}^{(i)})||p_0(\mathbf{z}_t|\mathbf{z}_{t-1}^{(\ell)})) \,,
\end{split}
\end{align}
%
%
where $\ell$ is the Monte Carlo sample index. The variational distribution is conditional on all time instants in $\mathbf{x}^{(i)}$ due to its implementation using an RNN, the coefficients of which may depend on all observations. The expression in~\eqref{eq:mc_approx_dkf} can be simplified by introducing the unnormalized weights as $w^{(i,\ell)} = w(\mathbf{x}^{(i)},\mathbf{z}^{(\ell)},{\boldsymbol{\theta}}) = p_{\boldsymbol{\theta}}(\mathbf{x}^{(i)},\mathbf{z}^{(\ell)})/q_{\boldsymbol{\phi}}(\mathbf{z}^{(\ell)}|\mathbf{x}^{(i)})$. An unbiased estimator of the DKF objective function can be calculated as
%
%
\begin{equation}
\label{eq:monte_carlo_vae}
    \nabla_{\boldsymbol{\theta}}\tilde{\mathcal{L}}^{\text{DKF}}({\boldsymbol{\theta}}, {\boldsymbol{\phi}}; \mathbf{x}^{(i)}) =  \frac{1}{L}\sum_{\ell=1}^{L}\nabla_{\boldsymbol{\theta}} \log w_{\text{DKF}}^{(i,\ell)} \,,
\end{equation}
where $w_{\text{DKF}}$ includes the objective function factorization as given by the DKF. 
%
%
%
%
%
%
Typically, when evaluating the expectations in the bound, a single sample ($L=1$) is drawn from the recognition network during learning, given the assumption of a sufficiently large minibatch size $N$. Accordingly, we adopt $L=1$ in this paper.


\subsection{Importance Weighted Autoencoders}
MCOs generalize the ELBO to objectives defined by taking the logarithm of a positive, unbiased estimator of the marginal likelihood as $\mathcal{L}(\mathbf{x}) = \mathbb{E}[\log\hat{p}_N(\mathbf{x})]$.
%
%
%
%
The tightness of an MCO is related to the variance of $\hat{p}_N(\mathbf{x})$~\cite{maddison_filtering_2017}. Consequently, multiple works proposed alternative objective functions that are tighter than the ELBO. 
We focus on the IWAE bound, derived from the $K$-sample importance weighting estimate of the marginal log-likelihood:
%
%
%
%
\begin{equation}
\label{eq:iwae_bound}
    \mathcal{L}_K^{\text{IWAE}}(\boldsymbol{\theta}, \boldsymbol{\phi}; \mathbf{x}) = \mathbb{E}_{\mathbf{z}^{(1)},\,\hdots,\,\mathbf{z}^{(K)}\sim\,q_{\boldsymbol{\phi}}(\mathbf{z}|\mathbf{x})}\left[\log\frac{1}{K}\sum_{k=1}^{K}w_{\text{VAE}}^{(k)}\right],
\end{equation}

\noindent where $\mathbf{z}^{(1)},\,\hdots,\,\mathbf{z}^{(K)}$ are sampled independently from the recognition model and $w_{\text{VAE}}$ considers the non-dynamical structure from a regular VAE~\cite{burda_importance_2016}. 
%
%
%
%
%


\subsection{IW-DKF Update Rule}\label{sec:iw-dkf}
It has been shown that the IWAE bound gets monotonically tighter for an increasing number of importance weighting samples~\cite[Appendix A]{burda_importance_2016}. This enables the learning of data representations closer to the true data distribution.
%
In the light of this, we incorporate sampling into the DVAE objective. $K$ samples drawn from the recognition network can be used to obtain an unbiased estimator of the IW-DKF objective gradients as 
\begin{equation}
\label{eq:monte_carlo_iw-dkf}
    \nabla_{\boldsymbol{\theta}}\tilde{\mathcal{L}}_K^{\text{IW-DKF}}({\boldsymbol{\theta}}, {\boldsymbol{\phi}}; \mathcal{D}^N) = \sum_{i=1}^{N}\sum_{k=1}^{K}\tilde{w}_{\text{DKF}}^{(i,k)}\nabla_{\boldsymbol{\theta}} \log w_{\text{DKF}}^{(i,k)} \,,
\end{equation}
where the contribution of all the datapoints in the minibatch $\mathcal{D}^N$ is considered and $\tilde{w}^{(i,k)} = {{w}^{(i,k)}}/{\sum_{k=1}^K{w}^{(i,k)}}$ are the normalized importance weights.
To compute the normalized importance weights, the marginal likelihood is evaluated as
\begin{equation}
\label{eq:ll_estimate}
    p_{\boldsymbol{\theta}}(\mathcal{D}^N  ) \approx \frac1{K}\sum^K_{k=1}\frac{p_{\boldsymbol{\theta}}(\mathcal{D}^N  |\mathbf{z}^{(k)})p_{\boldsymbol{\theta}}(\mathbf{z}^{(k)})}{q_{\boldsymbol{\phi}}(\mathbf{z}^{(k)}|\mathcal{D}^N )} \,,
\end{equation}
which is equivalent to a $K$-sample Monte-Carlo estimate of the likelihood. The log-sum-exp trick can be employed to compute this quantity in a numerically stable manner~\cite[Appendix A]{krishnan_structured_2016}.
%
In the implementation, gradients are aggregated across minibatches within the same epoch. For ease of exposition, the model sequential structure is not displayed. 
\label{subsec:proposed_method}
%

%% file: paper_sections/experiments.tex

We perform a twofold evaluation of the IW-DKF: first on the generative modeling performance using polyphonic data, and then in terms of state and parameter inference with the Lorenz attractor. 

\subsection{Learning DMMs with Polyphonic Music Data}

\subsubsection{Setup}
The DKF framework proposed in~\cite{krishnan_structured_2016} serves as our benchmark. 
For this experiment, most settings are consistent with the ones in~\cite[Polyphonic Music Experiment]{krishnan_structured_2016}. The inference network $q_{\boldsymbol\phi}(\mathbf{z}_t|\mathbf{z}_{t-1}, \mathbf{x})$ is fixed and includes an RNN and a combiner function. We learn a DMM on a polyphonic music dataset where an instance in the sequence $\mathbf{x}_t \in \mathbb{R}^{88}$ comprises an 88-dimensional binary vector corresponding to the notes of a piano.
Training, validation, and testing include 220, 76, and 77 sequences, respectively, with maximum lengths of 129, 144, and 160 polyphonic samples.
The KL divergence is annealed in the objective function from 0 to 1 over 5000 parameter updates. Results are reported based on early stopping using a validation set. An importance sampling-based estimate of the marginal log-likelihood is provided following~\eqref{eq:ll_estimate} considering the samples in a test set $\mathcal{D}_{\text{test}}$. 
The latent states in this dataset $\mathbf{z}_t \in \mathbb{R}^{100}$ do not have a physical meaning and consequently it is not possible to evaluate state estimation performance for this experiment. 
Sequential state and parameter estimation are evaluated in the next experiment.

\subsubsection{Results}

\begin{figure}[]
\centering
\vspace{-3ex}
\includegraphics[width=0.82\linewidth]{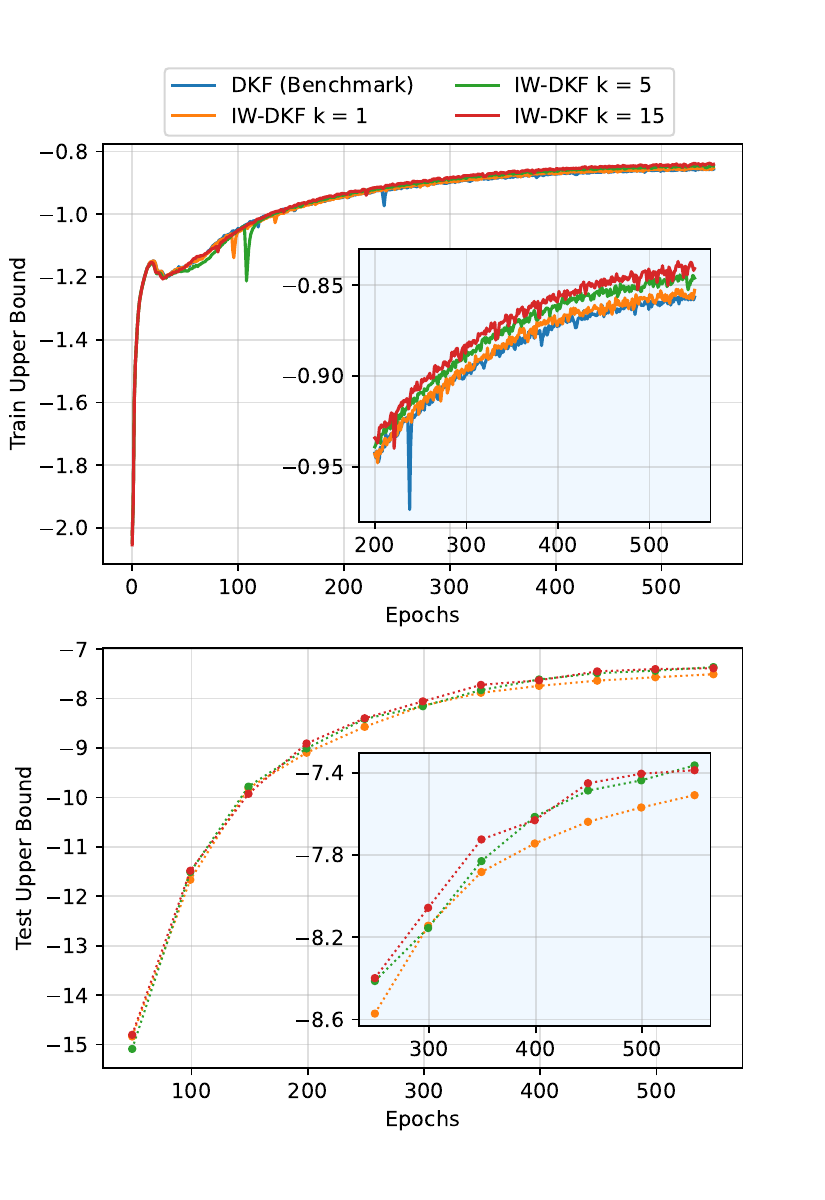}
\vspace{-4.5ex}
\caption{Training (upper plot) and test (lower plot) log-likelihood for the DKF~\cite{krishnan_deep_2015,krishnan_structured_2016} and the IW-DKF for $K \in\{ 1,\, 5,\, 15\}$ with polyphonic data.}
\label{fig:results_poly}
\end{figure}

The train and test log-likelihoods obtained with the DKF and the IW-DKF for $K \in\{ 1,\, 5,\, 15\}$, tested on polyphonic data, are shown in Figure~\ref{fig:results_poly}. Results obtained with the IW-DKF for $K = 1$ align with the DKF benchmark, which is expected as the VAE and IWAE update rules are equivalent for $K = 1$. To better observe the training convergence, particularly from epoch 200 onward, it is necessary to zoom in the figure. An improvement for $K > 1$ becomes evident, as the highest bound estimation is achieved for $K = 15$. Results for $K \in\{ 5,\, 15\}$ are very similar, especially with validation data. 
%
Table~\ref{table:results_poly} shows the mean and standard deviation of the training and validation log-likelihood and KL divergence for $K \in\{ 1,\, 5,\, 15\}$. For $K > 1$, the standard deviation of the estimated log-likelihood decreases from 0.029 to 0.008 in training and from 0.041 to 0.007 in validation, showing that an increasing number of samples provides a tighter log-likelihood estimate.
KL divergences between the variational distribution and the transition model decrease from 0.047 with $K=1$ to 0.045 with $K=5$ and 0.044 with $K =15$ in the validation stage. 
Overall, a consistent improvement in terms of generative modeling is observed in this experiment.


\begin{table}[]
\centering
\fontsize{8.3}{9.5}\selectfont
\caption{Mean and standard deviation (in parentheses) of the training and validation log-likelihood and KL divergence for the IW-DKF and $K \in\{ 1,\, 5,\, 15\}$ with polyphonic data. Results have been averaged across epochs after convergence.}
\vspace{-3ex}
\begin{tabular}{clllll}
\\ \hline
\multicolumn{1}{l}{} & \multicolumn{2}{c}{{ LL}}                                                          & \multicolumn{2}{c}{{ KL}}                               \\ \hline
k                    & \multicolumn{1}{c}{Train} & \multicolumn{1}{c}{Validation}  & \multicolumn{1}{c}{Train} & \multicolumn{1}{c}{Validation} \\ \hline
1                    & -0.865 (0.029)            & -0.888 (0.041)                       & 0.045 (0.006)             & 0.047 (0.012)                  \\ 
5                    & -0.853 (0.008)            & -0.876 (0.007)                          & 0.044 (0.002)             & 0.045 (0.002)                  \\ 
15                   & -0.848 (0.008)            & -0.875 (0.007)                          & 0.043 (0.002)             & 0.044 (0.002)                 \\ \hline
\end{tabular}
\label{table:results_poly}
\end{table}

\begin{figure*}
\centering
  \includegraphics[width=0.95\textwidth,height=3.6cm]{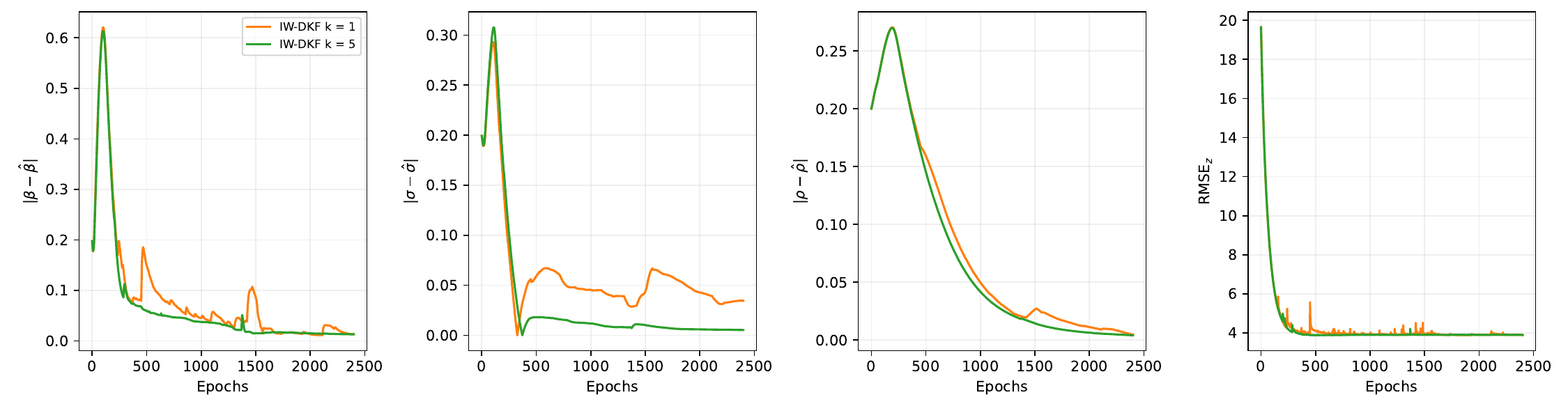}
  \vspace{-0.1cm}
  \caption{IW-DKF $\text{RMSE}_\mathbf{z}$ and error between estimated and true parameters with validation data as a function of training epochs for the 3-space Lorenz attractor model and $K \in\{ 1,\, 5\}$.}
   \label{fig:results_chaotic_params_states}
   \vspace{-2.5ex}
\end{figure*}

\subsection{State Estimation with a Physics-based Model}
%

\subsubsection{Setup}
We consider the 3-space Lorenz attractor~\cite{Lorenz1963DeterministicNF} as an example of non-linear chaotic system with known parametric form of the mean and covariance of the transition and emission models. This allows us to analyze the IW-DKF performance in terms of parameter and state estimation as in~\cite{2022_fusion_lorenz}. This model consists of three ordinary differential equations discretized as
\begin{align}
\begin{split}
&\mathbf{z}_t = \mathbf{z}_{t-1} + T_s f(\mathbf{z}_{t-1}) + \mathbf{q}_t \\
&\mathbf{x}_t = \mathbf{z}_t + \mathbf{r}_t,
\end{split}
\end{align}
where $\mathbf{z}_t \in \mathbb{R}^3$ and $\mathbf{x}_t \in \mathbb{R}^3$ are the latent state and observation at time instant $t$ and $\boldsymbol{\theta} = (\sigma,\, \rho,\, \beta)^\top$ are the generative model parameters. The vector valued function $\mathbf{f}(\cdot) = [f_1(\cdot), f_2(\cdot), f_3(\cdot)]^\top$ represents the dynamical model as $f_1(\mathbf{z}_{t})  = \sigma ({z}_{t,2} - {z}_{t,1})$, $f_2(\mathbf{z}_{t})  =  {z}_{t,1} (\rho - {z}_{t,3}) $ and $f_3(\mathbf{z}_{t})  = {z}_{t,1} {z}_{t,2} -  \beta {z}_{t,3}$.
%
%
%
In the experiments, the state and measurement noise are set to $\mathbf{r}_t,\, \mathbf{q}_t \sim \mathcal{N}(\mathbf{0}, 0.1\,\mathbf{I})$.
The initial state is $\mathbf{z}_0 = (1, 1, 1)^\top$, with generative model parameters $\sigma=28$, $\rho=10$, and $\beta=8/3$. Also, we set $T_s = 0.01$s with a total of  100 samples per sequence. Training includes 50 sequences, while validation and testing include 10 sequences. Only $K \in\{ 1,\, 5\}$ are tested given the similar results obtained for $K \in\{ 5,\, 15\}$ in the previous experiment.
%
\subsubsection{Results}
\begin{figure}[H]
\centering
\includegraphics[width=0.7\linewidth]{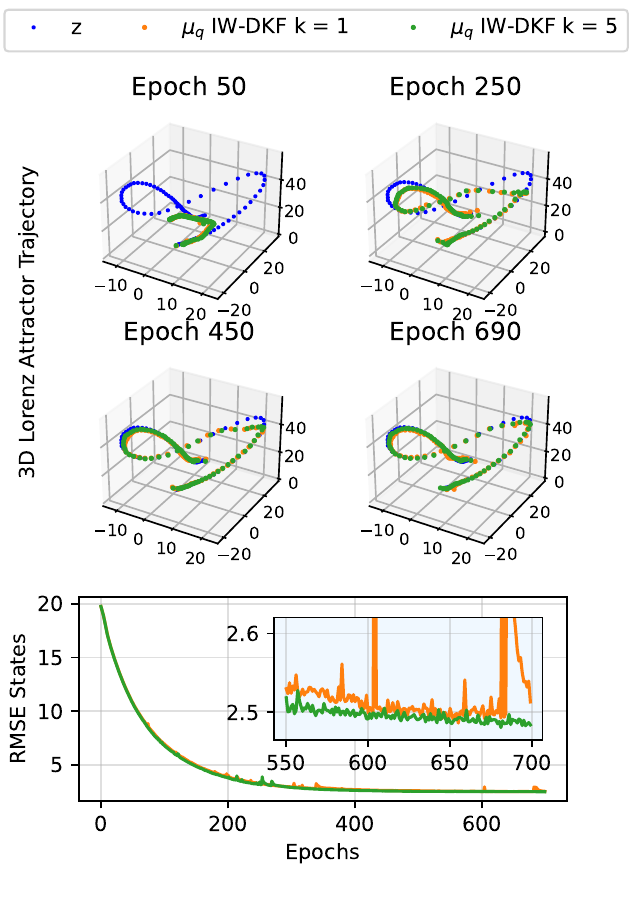}
\vspace{-2.5ex}
\caption{For $K \in\{ 1,\, 5\}$: (Top) 3-space Lorenz attractor trajectory (true and reconstructed by the IW-DKF) for one validation datapoint along training. (Bottom) IW-DKF state estimation RMSE with validation data.}
\label{fig:results_chaotic_oneMCrealization}
\vspace{0ex}
\end{figure}

Generative modeling results for this experiment may be found in Table~\ref{tab:results_chaotic_generative_modeling}. A tighter bound is obtained for $K = 5$, as the standard deviation decreases from $0.13$ to $0.02$ and $0.026$ in training and validation, respectively. The log-likelihood estimates increase from $-2.61$ with $K=1$ to $-1.94$ with $K=5$ in training and from $-2.63$ to $-1.92$ in validation. The variational distribution and the transition model also become closer for $K=5$, which can be understood from the decrease in KL divergence.
In Figure~\ref{fig:results_chaotic_oneMCrealization}, the IW-DKF reconstructed trajectories for one validation datapoint show successful results, especially at further epochs in time. As an improvement for $K = 5$ w.r.t. $K=1$ is not clear from the reconstructed trajectory plots, the state estimation RMSE is plotted as a function of the training epochs in the bottom subplot and also in Figure~\ref{fig:results_chaotic_params_states}. We define $\text{RMSE}_\mathbf{z} = \sqrt{\frac{1}{T}\sum^T_{t=1}\lVert \mathbf{z}_t-\hat{\mathbf{z}}_t\rVert^2}$, where the estimate $\hat{\mathbf{z}}_t$ is the mean of $q_{\boldsymbol{\phi}}(\mathbf{z}|\mathbf{x})$. This metric shows more stability and also lower values for $K=5$. The smoothing effect provided by the tighter IW-DKF bound can be observed in Figure~\ref{fig:results_chaotic_params_states} both in terms of parameter and state estimation. Results from Figure~\ref{fig:results_chaotic_params_states} are averaged after convergence, i.e., approximately epoch 600, and presented in Table~\ref{tab:results_chaotic_state_parameter}, showing a clear decrease in the error between the estimated and the true parameters for $K = 5$. Although the state estimation RMSE is only $0.016$ (unitless) lower for $K=5$, whether this improvement is notable or not depends on the application. However, given the chaotic structure of the data in this experiment, even the smallest change in a state value can cause completely different trajectories. Consequently, obtaining accurate state reconstructions can be very challenging.
%

\begin{table}[t]
\centering
\fontsize{8.3}{9.5}\selectfont
\caption{Mean and standard deviation (in parentheses) of the training and validation log-likelihood and KL divergence for the IW-DKF with the 3-space Lorenz attractor model. Results have been averaged across epochs after convergence.}
\vspace{-3ex}
\begin{tabular}{clllll}
\\ \hline
\multicolumn{1}{l}{} & \multicolumn{2}{c}{{ LL}}                                                          & \multicolumn{2}{c}{{ KL}}                               \\ \hline
$K$                    & \multicolumn{1}{c}{Train} & \multicolumn{1}{c}{Validation}  & \multicolumn{1}{c}{Train} & \multicolumn{1}{c}{Validation} \\ \hline
1  & -2.61 (0.13)     & -2.63 (0.13)          & 5.1 (1.05)      & 5.2 (0.96)                  \\ 
5    & -1.94 (0.02)   & -1.92(0.026)    & 4.64 (0.39)   & 4.34 (0.35)                  \\ \hline
\end{tabular}
\label{tab:results_chaotic_generative_modeling}
%
\bigskip
\fontsize{8.3}{9.5}\selectfont
\centering
\caption{IW-DKF $\text{RMSE}_\mathbf{z}$ and error between estimated and true parameters with the 3-space Lorenz attractor model. Results from Figure~\ref{fig:results_chaotic_params_states} have been averaged after convergence.}
\vspace{-0.7ex}
\begin{tabular}{ccccc}
\hline 
$K$ & \multicolumn{1}{c}{$|\beta-\hat{\beta}|$} & \multicolumn{1}{c}{$|\sigma-\hat{\sigma}|$} & \multicolumn{1}{c}{$|\rho-\hat{\rho}|$} & \multicolumn{1}{c}{$\text{RMSE}_\mathbf{z}$} \\\hline 
1 & 0.019                    & 0.035                     & 0.0085                  & 3.917                      \\
5 & 0.014                    & 0.005                     & 0.005                   & 3.901                    \\ \hline 
\end{tabular}
\vspace{-0.5cm}
\label{tab:results_chaotic_state_parameter}
\end{table}

%

%% file: paper_sections/conclusion.tex
In this paper, we introduced the IW-DKF by incorporating sampling into the DKF framework. We have investigated how a tighter bound capable of providing higher log-likelihoods on density estimation benchmarks can lead to more efficient state inference. An improvement in terms of generative modeling has been shown both when working with DMMs and a highly non-linear physics-based model: a 3-space Lorenz attractor.
Results suggest that applying sampling in the objective function provides more accurate and stable parameter and state estimates. A comparative study to determine which MCOs provide better results in terms of state inference is a subject for future research. 
Future work will also investigate the 
direct optimization of the variational distribution~\cite{finke2019importanceweighted} as an approach to improve state and parameter estimation in dynamical settings.
%